\documentclass{article}

\usepackage[english]{babel}
\usepackage[numbers, sectionbib, sort]{natbib}
\usepackage{float}

\usepackage[letterpaper,top=2cm,bottom=2cm,left=3cm,right=3cm,marginparwidth=1.75cm]{geometry}

\usepackage{amsmath}
\usepackage{graphicx}
\usepackage[colorlinks=true, allcolors=blue]{hyperref}

\title{\Large \bf Privacy Risks of Robot Vision: A User Study on Image Modalities and Resolution}
\author{Xuying Huang \quad Sicong Pan \quad Maren Bennewitz}

\begin{document}
\date{} 
\maketitle

\begingroup
\renewcommand{\thefootnote}{}
\footnotesize
\footnote{All authors are with the Humanoid Robots Lab, University of Bonn, Germany. Maren Bennewitz is additionally with the Lamarr Institute for Machine Learning and Artificial Intelligence, Bonn, Germany. This work has been partially funded by the Privatar project, grant No.16KIS1949. Corresponding: \texttt{xhuang1@uni-bonn.de}.
}
\endgroup

\begin{abstract}
User privacy is a crucial concern in robotic applications, especially when mobile service robots are deployed in personal or sensitive environments.
However, many robotic downstream tasks require the use of cameras, which may raise privacy risks.
To better understand user perceptions of privacy in relation to visual data, we conducted a user study investigating how different image modalities and image resolutions affect users' privacy concerns.
The results show that \textbf{depth} images are broadly viewed as privacy-safe, and a similarly high proportion of respondents feel the same about \textbf{semantic} segmentation images.
Additionally, the majority of participants consider 32×32 resolution \textbf{RGB} images to be almost sufficiently privacy-preserving, while most believe that 16×16 resolution can fully guarantee privacy protection.
\end{abstract}

\section{Introduction}

With the rapid advancements in robotics, mobile service robots have become increasingly essential in assisting people with a wide variety of tasks, including domestic chores, healthcare, and package delivery~\citep{cakmak2013towards, asgharian2022review, alverhed2024autonomous}.
To efficiently accomplish these tasks, most mobile robots are equipped with high-resolution cameras that capture detailed visual data of their operational environments.
Although the usage of these visual sensors enhances robot performance, it simultaneously raises substantial privacy concerns~\citep{nieto2024robot}, particularly when robots operate within users' personal or private spaces. 

Motivated by the critical need to balance robot performance and user privacy, it is important to understand user perceptions of privacy concerns related to robotic visual data collection.
Previous studies have investigated privacy concerns related to general camera surveillance~\citep{huang2023rethinking, el2024through}, yet relatively few studies specifically focus on visual data modalities and image resolution for privacy in the context of mobile robotics.
Therefore, we conducted a user study aimed at uncovering user preferences and attitudes regarding privacy risks associated with robotic visual perception.
Specifically, our objectives include evaluating user opinions on privacy of different visual data modalities and determining user-preferred strategies and thresholds (e.g., reduced image resolution) for effective privacy preservation. 

\section{Methodology}

\subsection{Study Design}

To explore the impact of robotic visual data on user privacy, we designed a targeted survey to address the following key questions:
\begin{itemize}
    \item \textbf{Attitudes toward full visibility}: Are users comfortable with robots having unrestricted visual access to all areas within their private spaces?
    \item \textbf{Sensitivity of image modalities}: How do users perceive different image types (e.g., RGBs, depths, semantics) in terms of privacy intrusion? Which modalities are considered acceptable?
    \item \textbf{Preferences for privacy-preserving image processing}: Do users prefer high-resolution images subsequently blurred or inherently low-resolution images captured directly by the camera?
    \item \textbf{Resolution thresholds for privacy protection}: When using inherently low-resolution RGB image capture as a privacy-preserving approach, what resolution levels do users consider sufficient to feel their privacy is protected?
\end{itemize}

\subsection{Survey Platform}
\setcounter{footnote}{0}
The survey was conducted using the \textit{FreeOnlineSurveys}\footnote{\url{https://app.freeonlinesurveys.com/}} platform, a widely-used tool for designing and distributing web-based questionnaires.
The platform allows us to create a structured and interactive survey, incorporating single-choice, multiple-choice, open-ended questions, along with embedded visual content to capture participants' perceptions regarding visual data and privacy for mobile service robots.

Participant recruitment was carried out through public online platforms, including social media and mailing lists.
Respondents were invited to complete the survey anonymously, ensuring the protection of their personal information. 

\section{Survey Results}

The survey involved a total of 62 participants(mean age 27 years-old; 69\% female, 31\% male) and provided several important insights into user privacy preferences.
\begin{itemize}
    \item \textbf{Attitudes toward full visibility}: In this user study, participants were asked: “Do you allow mobile service robots to scan everything clearly in your private spaces (e.g., your room, office)?” As shown in Figure~\ref{willingness}, a majority of respondents (66\%) indicate that they would not allow mobile service robots to scan everything clearly within their private spaces, such as bedrooms or offices. This result highlights a strong user preference for privacy protection and suggests a general discomfort with unrestricted visual access by robots in personal environments.
    
    \begin{figure}[H]
    \centering
    \includegraphics[width=0.7\textwidth]{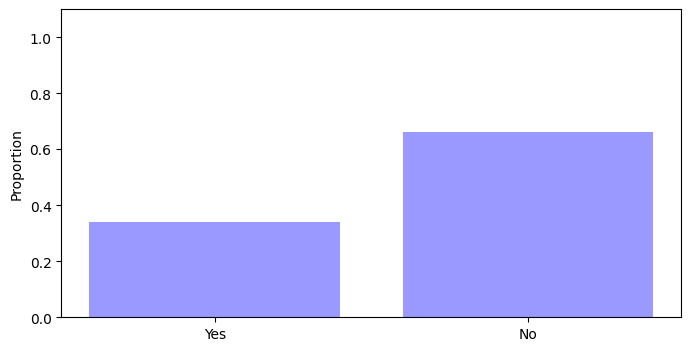}
    \caption{
    Willingness to Allow Robots to Scan Private Spaces. 
    } 
    \label{willingness}
    \end{figure}
    
    \item \textbf{Sensitivity of image modalities}: Participants were shown the same scene in three different image modalities (RGB, depth, and semantic segmentation), as depicted in Figure.~\ref{modality_case}, and asked two questions to investigate privacy situation on these different modalities: “For the images below, which type of image (image modality) do you believe could violate your privacy or still reveal private information? (Multi Select)” and “For the images below, which type of image (image modality) do you think would preserve your private information? (Multi Select)”.
    
    Approximately 77\% of participants believe that RGB images pose a threat to their privacy, as illustrated in Figure~\ref{modality_risk}. This indicates a strong association between high-fidelity, color-rich visual data and potential privacy violations, likely due to the fact that RGB images can clearly reveal identifiable details such as faces, personal information, or sensitive environments. In contrast, Figure~\ref{modality_preserve} shows that 67\% of users consider depth images to be non-intrusive, and 52\% feel the same about semantic segmentation images. These modalities abstract away specific visual features and present scenes in a more symbolic or structural form, which may help users feel that their identities or private information are less likely to be exposed.

    \begin{figure}[H]
    \centering
    \includegraphics[width=0.8\textwidth]{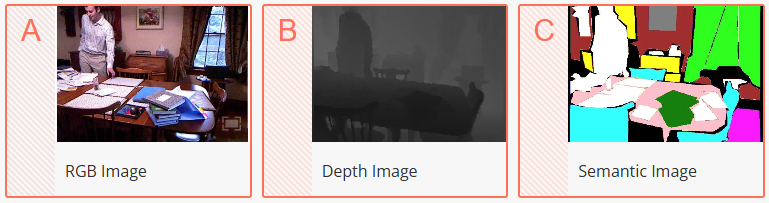}
    \caption{
    Same Scene with Different Image Modality.
    } 
    \label{modality_case}
    \end{figure}

    \begin{figure}[H]
    \centering
    \includegraphics[width=0.7\textwidth]{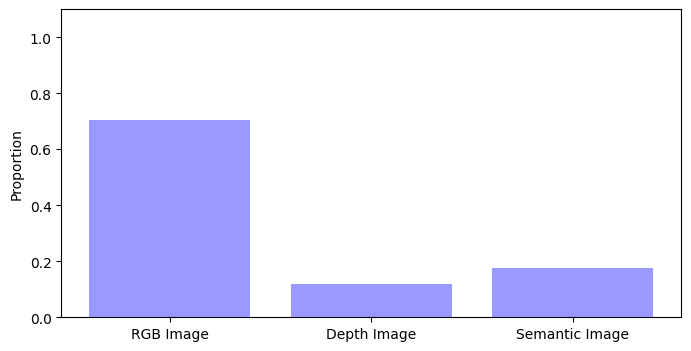}
    \caption{
    Perceived Privacy Risks by Image Modality.
    } 
    \label{modality_risk}
    \end{figure}

    \begin{figure}[H]
    \centering
    \includegraphics[width=0.7\textwidth]{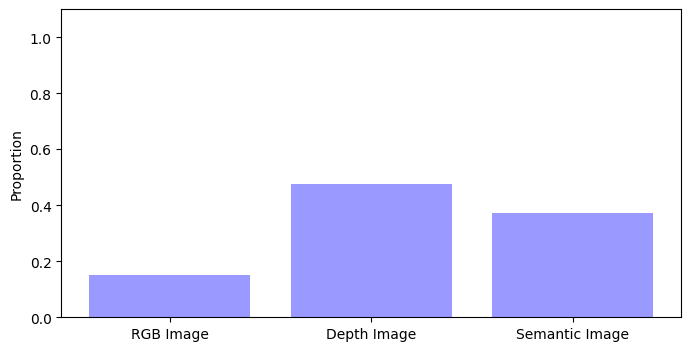}
    \caption{
    Perceived Privacy Preservation by Image Modality.
    } 
    \label{modality_preserve}
    \end{figure}

    \item \textbf{Preferences for privacy-preserving image processing}: In order to investigate users' preferences on how to process images to protect privacy, we asked a question "For protecting your private information, how would you prefer the mobile service robots' camera to process the data?" Figure.~\ref{method_case} shows an example of two different ways to process images.
    
    Most users favor capturing images directly at low resolutions rather than initially acquiring high-resolution images followed by blurring, as demonstrated in Figure~\ref{method}. This preference suggests that users are more comfortable with minimizing data collection at the source, possibly due to a perception that once high-resolution data is captured, there is an inherent risk of privacy breach, even if post-processing is applied later. The results highlight a user tendency to favor preventive over corrective measures in privacy protection, emphasizing the importance of designing privacy-aware vision systems that limit sensitive information from being recorded in the first place.

    \begin{figure}[H]
    \centering
    \includegraphics[width=0.7\textwidth]{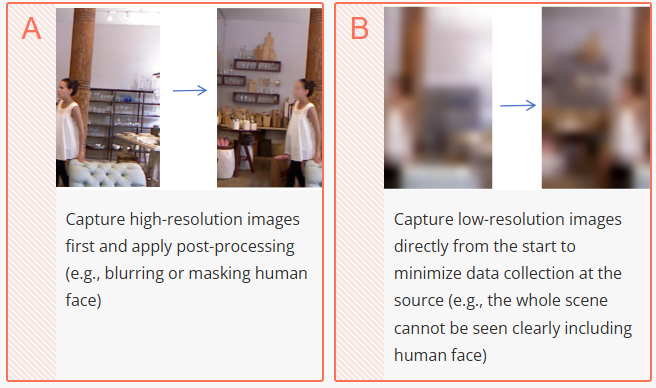}
    \caption{
    Two Different Camera Data Processing Methods for Privacy Protection.
    } 
    \label{method_case}
    \end{figure}

    \begin{figure}[H]
    \centering
    \includegraphics[width=0.7\textwidth]{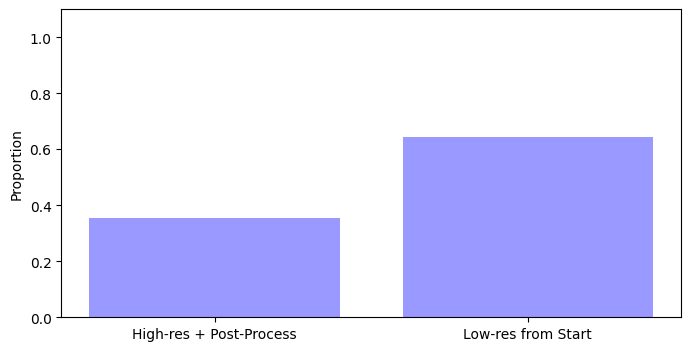}
    \caption{
    Preferred Camera Data Processing Methods for Privacy Protection.
    } 
    \label{method}
    \end{figure}
    
    \item \textbf{Resolution thresholds for privacy protection}: To better understand users’ preferences regarding privacy-preserving RGB image resolutions, the survey presented participants with a series of progressively downsampled images on two cases (see Figure~\ref{case} \& Figure~\ref{case_2}) and asked two key questions: (1) "If the camera uses a low-resolution setting directly upon activation, what resolution level would you prefer for desired privacy? (resolution you think is okay to meet privacy protection needs)" and (2) "If the camera uses a low-resolution setting directly upon activation, what resolution level would you prefer to ensure privacy? (resolution you think is necessary to fully guarantee privacy protection) "
    
    As shown in Figure~\ref{basic_1} \& Figure~\ref{basic_2}, when asked about the minimum resolution that would be acceptable to meet basic privacy needs, the most popular choice is $32 \times $32, selected by 29.0\% and 38.7\% of respondents for case 1 and case 2, respectively. This indicates that a large portion of users believe the resolution must be reduced to a relatively low level, such as $32 \times $32, in order to protect their basic privacy needs. 
    
    While asked about the resolution necessary to fully guarantee privacy as shown in Figure~\ref{necessary_1} and Figure~\ref{necessary_2}, responses shift toward lower resolutions. Here, $16 \times $16 becomes the most selected option in both scenarios, with 27.4\% in the first case and 29\% in the second, considering it the threshold for ensuring complete privacy. Whereas $32 \times $32 is selected by only 17.7 \% and 24\% of respondents respectively. These changes suggest that although this resolution may be acceptable for general privacy needs, it is not perceived as sufficient by those with stronger privacy concerns.  

    \begin{figure}[H]
    \centering
    \includegraphics[width=0.7\textwidth]{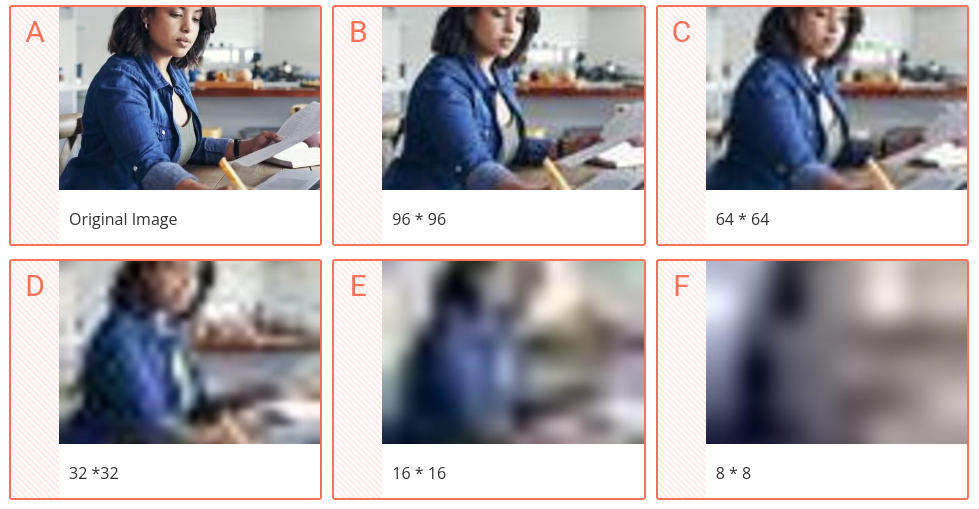}
    \caption{
    Case 1 with Six Resolutions for Basic vs. Full Privacy Requirements.
    } 
    \label{case}
    \end{figure}
    
    \begin{figure}[H]
    \centering
    \includegraphics[width=0.7\textwidth]{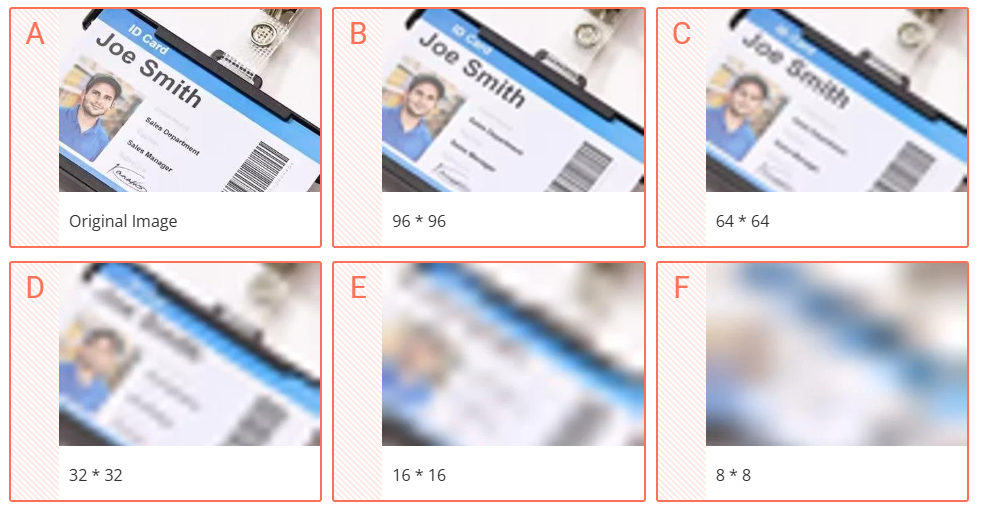}
    \caption{
    Case 2 with Six Resolutions for Basic vs. Full Privacy Requirements.
    } 
    \label{case_2}
    \end{figure}

    \begin{figure}[H]
    \centering
    \includegraphics[width=0.7\textwidth]{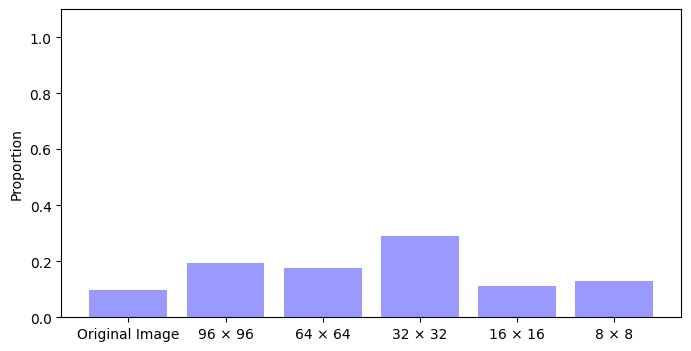}
    \caption{
    Acceptable Resolution for Basic Privacy Need for Case 1.
    } 
    \label{basic_1}
    \end{figure}

    \begin{figure}[H]
    \centering
    \includegraphics[width=0.7\textwidth]{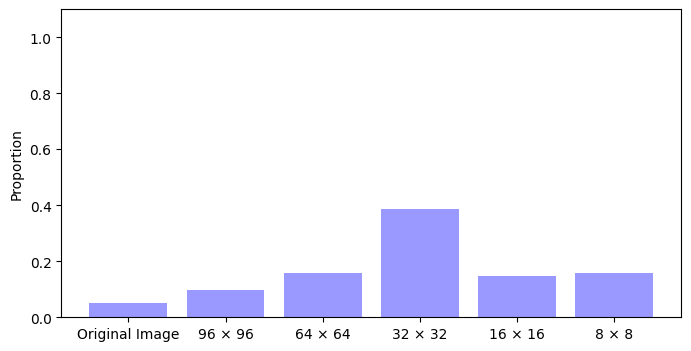}
    \caption{
    Acceptable Resolution for Basic Privacy Need for Case 2.
    } 
    \label{basic_2}
    \end{figure}

    \begin{figure}[H]
    \centering
    \includegraphics[width=0.7\textwidth]{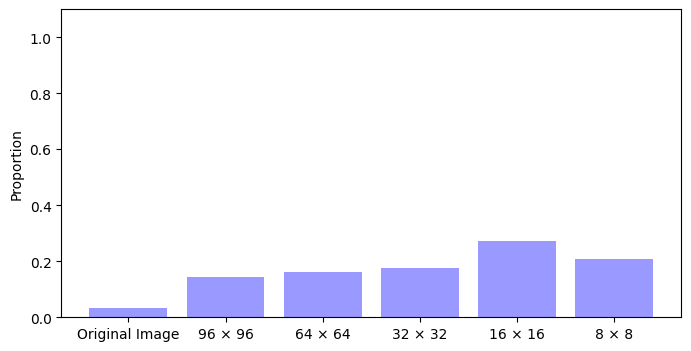}
    \caption{
    Preferred Minimum Resolution for Privacy Guarantee for Case 1.
    } 
    \label{necessary_1}
    \end{figure}
    
    \begin{figure}[H]
    \centering
    \includegraphics[width=0.7\textwidth]{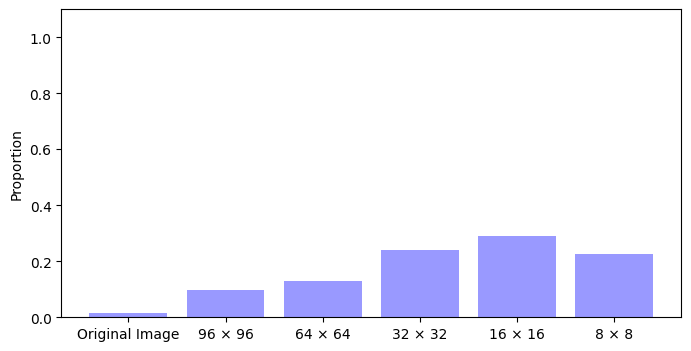}
    \caption{
    Preferred Minimum Resolution for Privacy Guarantee for Case 2.
    } 
    \label{necessary_2}
    \end{figure}

\end{itemize}

\section{Conclusions}

Through this user study, we explore individuals’ perceptions of how visual data captured by mobile service robots operating in private environments impacts their sense of privacy. 
The findings reveal a clear concern among users regarding visual access to personal spaces.
Most users are uncomfortable with robots having clear visibility into private spaces, due to fears of personal information exposure.
RGB images are viewed as the most privacy-invasive, while depth and semantic segmentation images are considered more privacy-friendly by the majority of respondents.
When it comes to privacy-preserving camera strategies, users prefer cameras that capture low-resolution images from the start, rather than relying on post-processing high-resolution data.
In terms of RGB image resolution, $32 \times $32 resolution is seen as sufficient for basic privacy protection, while $16 \times $16 resolution is preferred for stronger privacy assurance.
Taken together, these insights offer valuable guidance for the design of privacy-aware vision systems in mobile robotics, emphasizing the importance of early-stage data minimization and the careful selection of image modalities and resolutions.


\bibliographystyle{IEEEtranSN}
\footnotesize
\bibliography{tr}

\end{document}